\documentclass[9pt]{article}
\usepackage{spconf,amsmath,graphicx,hyperref}
\usepackage{amssymb}
\usepackage{booktabs}
\usepackage{multirow}
\usepackage{array}

\title{SFGNET: SEMANTIC AND FREQUENCY GUIDED NETWORK FOR CAMOUFLAGED OBJECT DETECTION}
\name{Dezhen Wang$^{1}$, Haixiang Zhao$^{1}$, Xiang Shen$^{2}$, Sheng Miao$^{1,*}$
	\thanks{$^{*}$Sheng Miao is the corresponding author.\\
	Emails~of~each~author: 202323050920@stu.qut.edu.cn,\\ 202523051002@stu.qut.edu.cn, xiangshen@gwu.edu, smiao@qut.edu.cn}}

\address{$^{1}$Qingdao University of Technology, Qingdao, Shandong, China\\
	$^{2}$The George Washington University, Washington, DC, USA}

\begin{document}
%
\maketitle
\begin{abstract}
Camouflaged object detection (COD) aims to segment objects that blend into their surroundings. However, most existing studies overlook the semantic differences among textual prompts of different targets as well as fine-grained frequency features. In this work, we propose a novel Semantic and Frequency Guided Network (SFGNet), which incorporates semantic prompts and frequency-domain features to capture camouflaged objects and improve boundary perception. We further design Multi-Band Fourier Module(MBFM) to enhance the ability of the network in handling complex backgrounds and blurred boundaries. In addition, we design an Interactive Structure Enhancement Block (ISEB) to ensure structural integrity and boundary details in the predictions. Extensive experiments conducted on three COD benchmark datasets demonstrate that our method significantly outperforms state-of-the-art approaches. The code will be released once the paper is accepted.
\end{abstract}
\vspace{-1em}
\section{Introduction}
\label{sec:intro}
\vspace{-0.5em}
In nature, many animals and plants blend into their surroundings by altering their color, shape, or other features to avoid predators, a survival strategy known as camouflage \cite{Cuthill2005,Fan2020}. Camouflaged object detection (COD) aims to segment these concealed objects. COD is essential for various applications, including medical image segmentation \cite{Fan2020}, industrial defect detection \cite{Bhajantri2006}, and agricultural pest monitoring \cite{Rustia2020}. CNN-based COD has achieved remarkable progress. However, these methods primarily capture local features while struggling with global contextual dependencies and often neglect fine-grained frequency-domain features, making detection in complex backgrounds challenging. To address these challenges, Cong et al. adopted a two-stage model for coarse localization and detail refinement \cite{Cong2023}; Yin et al. introduced masked separable attention with PVT backbone to effectively model global dependencies \cite{Yin2024}; Lou et al. proposed VSCode \cite{Luo2023} to improve segmentation performance under complex backgrounds; and Zhang et al. applied class-guided COD to enhance accuracy \cite{Zhang2024}. Although camouflaged objects resemble their environment in shape, color, and texture, they preserve significant semantic differences across categories. However, existing prompt-based methods fail to fully exploit the relationship between camouflaged objects and their background, and most studies overlook fine-grained frequency-domain features, leading to poor performance in scenarios with cluttered backgrounds, blurred boundaries, or multiple targets.
\begin{figure}[tb]
	\centerline{\includegraphics[width=\linewidth]{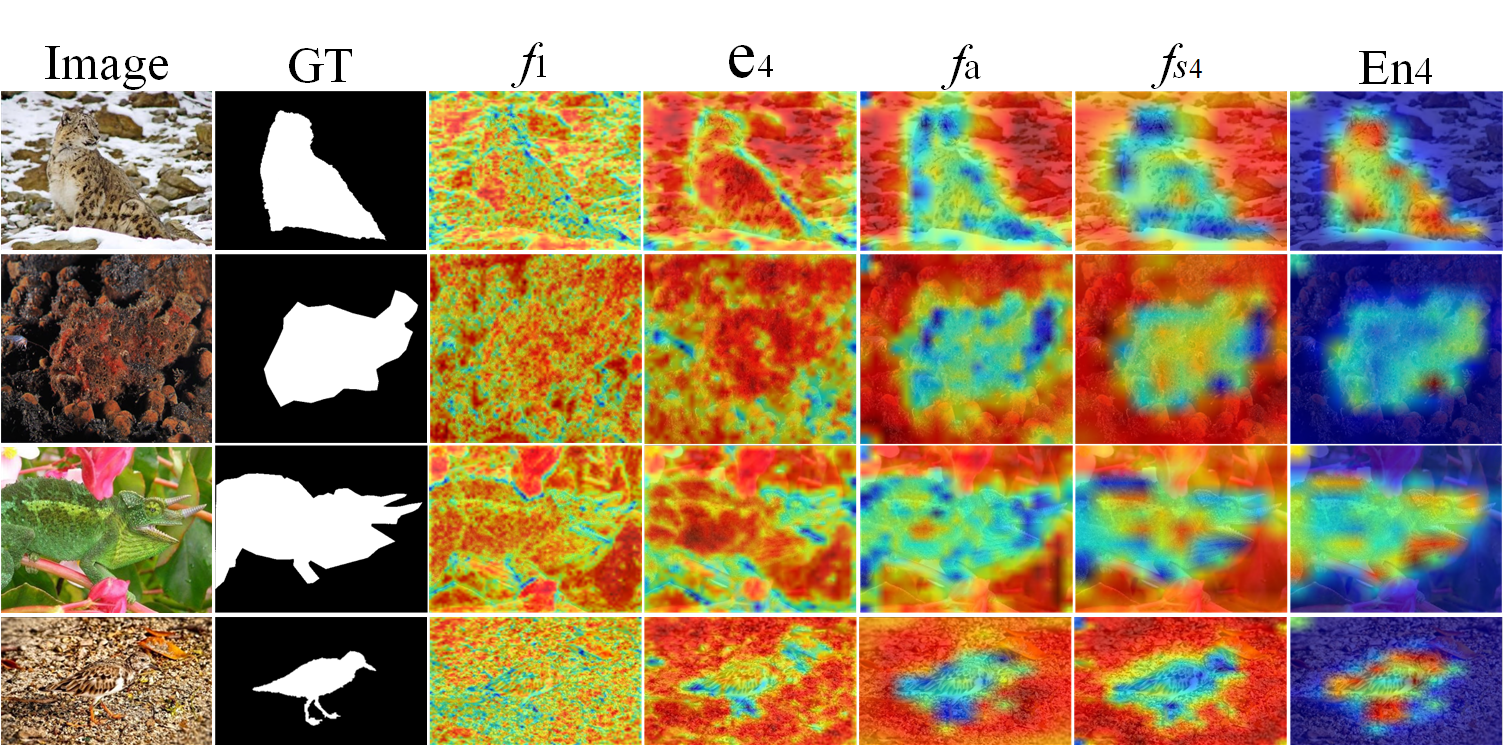}}
	\caption{Visualization of different intermediate feature perceptions in SFGNet.}
	\label{fig1}
\end{figure}

To address these challenges, we propose a novel Semantic and Frequency Guided Network (SFGNet), which enhances segmentation accuracy by integrating prompts that capture the relationship between camouflaged objects and their background with multi-band frequency representations. Fig. \ref{fig1} illustrates the intermediate feature representations of SFGNet. Specifically, we first integrate visual features and textual semantics through cross-modal interaction to enhance high-level semantic perception of camouflaged objects. Second, we introduce multi-band frequency modeling and frequency-spatial fusion to better capture fine-grained structures and boundaries. Finally, we design a novel Interactive Structure Enhancement Block (ISEB) to further improve the overall consistency and precision of predictions. 
\begin{figure*}[tb]
	\centerline{\includegraphics[width=\linewidth]{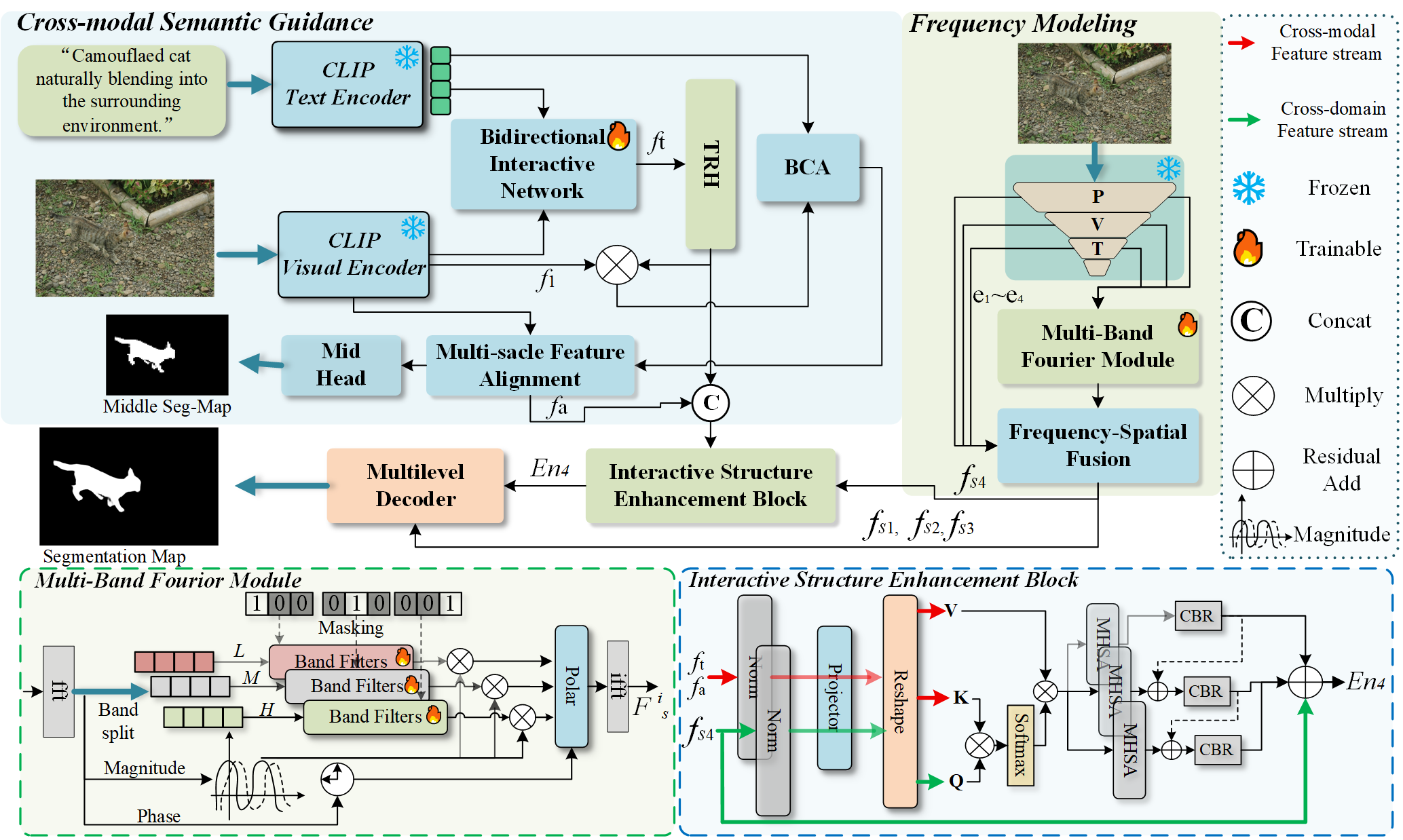}}
	\caption{Overall framework of SFGNet, which leverages textual semantics and frequency-domain features to guide the representation and segmentation of camouflaged objects. }
	\label{fig2}
\end{figure*}
The main contributions are summarized as follows:
\vspace{-0.5em}
\begin{itemize}
\item We introduce prompts that explicitly model the relationship between the target and the background, which effectively enhance the robustness of segmentation.
\vspace{-0.5em}
\item We incorporate multi-band frequency feature guidance to capture structural information across low, mid, and high frequencies, improving modeling of complex backgrounds and unclear boundaries.
\vspace{-0.5em}
\item We design cross-modal semantic interaction and structure enhancement modules, enabling effective fusion of semantic and visual features while ensuring structural integrity and boundary accuracy.
\end{itemize}
\vspace{-2em}
\section{Methodology}
\label{sec:methodology}
\vspace{-0.5em}
\subsection{Overview}
\vspace{-0.2em}
Fig. \ref{fig2} illustrates the overall framework of the proposed Semantic and Frequency Guided Network (SFGNet), which consists of Cross-modal Semantic Guidance and Frequency Modeling, where CLIP extracts semantic representations and the Pyramid Vision Transformer (PVT) provides multi-scale spatial features. The cross-modal features are enhanced via the Bidirectional Interaction Network (BIN), Bidirectional Cross-Attention (BCA), and Multi-Scale Feature Alignment (MFA), while the Multi-Band Fourier Module (MBFM) and Frequency-Spatial Fusion (FSF) integrate frequency- and spatial-domain information. Finally, the Interactive Structure Enhancement Block (ISEB) and a decoder progressively refine fused features to reconstruct accurate camouflaged object segmentation maps.
\vspace{-1em}
\subsection{Cross-modal Semantic Guidance}
\vspace{-0.2em}
The Cross-modal Semantic Guidance (CSG) module consists of the Bidirectional Interaction Network (BIN), Bidirectional Cross-Attention (BCA), and Multi-Scale Feature Alignment (MFA), which jointly achieve multi-scale interaction and cross-modal fusion. In BIN, the text feature $t$ generates gating coefficients to modulate multi-scale visual features $v_3, v_4, v_5$, as shown in Eq.~\ref{eq:bin_gate}. 
\vspace{-0.2em} 
\begin{equation}
	\hat{t} = \sigma(W_g t), \qquad 
	\hat{v}_i = W_i v_i \odot \hat{t}_i, \; i \in \{3,4,5\},
	\label{eq:bin_gate}
\end{equation}

BIN further refines features via top-down propagation and bottom-up feedback, as expressed in Eq.~\ref{eq:bin_flow}. 
\vspace{-0.2em}
\begin{equation}
	\begin{aligned}
		p_i     &= f(\hat{v}_i + \mathrm{Up}(p_{i+1})), \\
		n_{i+1} &= f(p_{i+1} + \mathrm{Down}(n_i)), \quad i \in \{3,4\},
	\end{aligned}
	\label{eq:bin_flow}
\end{equation}

The Bidirectional Cross-Attention (BCA) module strengthens cross-modal interaction by combining text-guided visual features and visual-guided textual features, as formulated in Eq.~\ref{eq:bca}. 
\vspace{-0.2em}
\begin{equation}
	F_{\mathrm{BCA}} = \alpha \cdot \hat{X}_t + \beta \cdot \mathrm{Expand}(\hat{T}_x),
	\label{eq:bca}
\end{equation}

The Multi-Scale Feature Alignment (MFA) module aligns visual features across scales to reduce semantic inconsistency, with the aligned representation given in Eq.~\ref{eq:mfa}. 
\vspace{-0.2em}
\begin{equation}
	F_{\mathrm{MFA}} = W_f \left[ F_1' \,\|\, F_2' \,\|\, F_3' \right],
	\label{eq:mfa}
\end{equation}

Here, $\|$ denotes channel concatenation and $W_f$ is a projection layer. Together, BIN, BCA, and MFA provide consistent and semantically enriched multi-scale features for subsequent modules.
\vspace{-1.4em}
\subsection{Frequency Modeling}
\vspace{-0.4em}
The Multi-Band Fourier Module (MBFM) enhances feature representation by mapping features into the frequency domain and decomposing them into multiple bands through a masking mechanism, as defined in Eq.~\ref{eq:mbfm_mask}.
\vspace{-0.5em}
\begin{equation}
	\tilde{M}_i(u,v) =
	\begin{cases}
		M(u,v), & (u,v) \in band_i, \\
		0, & \text{otherwise},
	\end{cases}
	\label{eq:mbfm_mask}
\end{equation}

Each band-specific component is fused with its phase and transformed back via inverse FFT, as shown in Eq.~\ref{eq:mbfm_ifft}.
\vspace{-0.5em}
\begin{equation}
	X_i = F^{-1}\!\left( \tilde{M}_i e^{jP} \right),
	\label{eq:mbfm_ifft}
\end{equation}

Finally, all band-specific features are concatenated and projected to yield the frequency-enhanced representation $F_{\mathrm{freq}}$. The Frequency-Spatial Fusion (FSF) module further integrates frequency- and spatial-domain features through complementary attention mechanisms. Channel attention is applied to spatial features, while average and max pooling on frequency features generate a spatial attention map, as formulated in Eq.~\ref{eq:fsf_simplified}.
\vspace{-0.5em}
\begin{equation}
	\begin{aligned}
		F_c &= F_{spa} \cdot \sigma\!\left(W_c(\mathrm{GAP}(F_{spa}))\right), \\
		F_s &= F_{freq} \cdot \sigma\!\left(f_s([\mathrm{Avg}(F_{freq}), \mathrm{Max}(F_{freq})])\right),
	\end{aligned}
	\label{eq:fsf_simplified}
\end{equation}

Here, $\sigma(\cdot)$ is the Sigmoid function and $[\cdot]$ indicates concatenation. The two branches are then merged to obtain the fused representation, as shown in Eq.~\ref{eq:fsf_fusion}.
\vspace{-0.5em}
\begin{equation}
	F_{fs} = \phi\!\left([F_c \,\|\, F_s]\right),
	\label{eq:fsf_fusion}
\end{equation}
\vspace{-2.4em}
\subsection{Interactive Structure Enhancement}
\vspace{-0.4em}
The Interactive Structure Enhancement Block (ISEB) refines deep features by attending to auxiliary features, enabling more consistent structures and clearer boundaries, as formulated in Eq.~\ref{eq:iseb}.
\vspace{-0.4em}
\begin{equation}
	\hat{F} = \mathrm{Softmax}\!\left(\tfrac{QK^{T}}{\sqrt{d}}\right)V,
	\label{eq:iseb}
\end{equation}

Here, $d$ is the channel scaling factor, and the output $\hat{F}$ is fused with the main feature through a residual connection.
\vspace{-1em}
\subsection{Loss Function}
\vspace{-0.4em}
To jointly ensure pixel-wise accuracy, regional consistency, and cross-modal alignment, we adopt a composite loss consisting of Weighted BCE, Weighted IoU, and Cosine Similarity, as summarized in Eq.~\ref{eq:loss}.
\vspace{-0.5em}
\begin{equation}
	L = L_{wbce} + L_{wIoU} + \lambda L_{cos},
	\label{eq:loss}
\end{equation}
\vspace{-3em}
\section{Experiments and Results}
\label{sec:pagestyle}
\vspace{-0.8em}
\subsection{Experiments}
\subsubsection{Datasets}
Our experiments were conducted on three publicly available datasets: CAMO \cite{Le2019}, COD10K \cite{Fan2021}, and NC4K \cite{Lv2021}. Following the standard training protocol, the training set consists of 1,000 images from CAMO and 3,040 images from COD10K, resulting in a total of 4,040 training images. The test set includes 250 images from CAMO, 2,026 from COD10K, and 4,121 from NC4K, yielding 6,397 images in total for evaluation.
{\renewcommand{\arraystretch}{0.85} 
\begin{table*}[tb]
	\centering
	\caption{Comparative experimental results of LMSGNet and state-of-the-art (SOTA) models on the CAMO, COD10K, and NC4K datasets. The best and second-best values in each column are denoted by bold and underline, respectively.}
	\resizebox{\textwidth}{!}{
		\begin{tabular}{l|l|l|cccc|cccc|cccc}
			\toprule
			\multirow{2}{*}{\textbf{Methods}} & \multirow{2}{*}{\textbf{Prompt}} & \multirow{2}{*}{\textbf{Backbone}} 
			& \multicolumn{4}{c|}{\textbf{CAMO (250 images)}} 
			& \multicolumn{4}{c|}{\textbf{COD10K (2,026 images)}} 
			& \multicolumn{4}{c}{\textbf{NC4K (4,121 images)}} \\
			\cmidrule(lr){4-7} \cmidrule(lr){8-11} \cmidrule(lr){12-15}
			& & & $S_m\uparrow$ & $F_\beta^\omega\uparrow$ & MAE$\downarrow$ & $E_m\uparrow$
			& $S_m\uparrow$ & $F_\beta^\omega\uparrow$ & MAE$\downarrow$ & $E_m\uparrow$
			& $S_m\uparrow$ & $F_\beta^\omega\uparrow$ & MAE$\downarrow$ & $E_m\uparrow$ \\
			\midrule
			PFNet \cite{Mei2021}      & None      & ResNet        & 0.782 & 0.695 & 0.085 & 0.855 & 0.800 & 0.660 & 0.040 & 0.877 & 0.829 & 0.745 & 0.053 & 0.887 \\
			ZoomNet \cite{Pang2020}    & None      & ResNet        & 0.820 & 0.752 & 0.066 & 0.877 & 0.838 & 0.729 & 0.029 & 0.888 & 0.853 & 0.784 & 0.043 & 0.896 \\
			FSPNet \cite{Huang2023}     & None      & ViT           & 0.851 & 0.802 & 0.056 & 0.905 & 0.850 & 0.755 & 0.028 & 0.912 & 0.879 & 0.816 & 0.035 & 0.914 \\
			FEDER \cite{He2023}      & None      & Res2Net       & 0.836 & 0.807 & 0.066 & 0.897 & 0.844 & 0.748 & 0.029 & 0.911 & 0.862 & 0.824 & 0.042 & 0.913 \\
			HitNet \cite{Hu2023}     & None      & PVT           & 0.849 & 0.809 & 0.055 & 0.906 & 0.871 & 0.806 & 0.023 & 0.935 & 0.875 & 0.834 & 0.037 & 0.926 \\
			FSEL \cite{Sun2022}       & None      & PVT           & 0.885 & 0.857 & 0.040 & 0.942 & 0.877 & 0.799 & 0.021 & 0.937 & 0.892 & 0.852 & 0.030 & 0.941 \\
			CamoFormer \cite{Yin2024} & None      & PVT           & 0.872 & 0.831 & 0.046 & 0.929 & 0.869 & 0.786 & 0.023 & 0.932 & 0.892 & 0.847 & 0.030 & 0.939 \\
			FHCNet \cite{Zhao2025}     & None      & Swin+Res2Net  & 0.885 & 0.842 & 0.039 & 0.934 & 0.869 & 0.778 & 0.023 & 0.931 & 0.893 & 0.843 & 0.031 & 0.937 \\
			VSCode \cite{Luo2023}     & 2D Prompt & Swin          & 0.873 & 0.820 & 0.046 & 0.925 & 0.869 & 0.780 & 0.025 & 0.931 & 0.882 & 0.841 & 0.032 & 0.935 \\
			CGCOD \cite{Zhang2024}      & Camoclass & CLIP+PVT & 0.896 & 0.874 & 0.036 & 0.947 & \underline{0.890} & 0.824 & \textbf{0.018} & \underline{0.948} & 0.904 & 0.879 & 0.026 & 0.949 \\
			\midrule
			SFGNet     & Semantics & CLIP+PVTv2-b2 & \underline{0.904} & 0.878 & 0.034 & \underline{0.949} & \underline{0.890} & 0.826 & \underline{0.019} & 0.947 & 0.907 & 0.879 & 0.025 & 0.952 \\
			SFGNet     & Semantics & CLIP+PVTv2-b4 & \textbf{0.906} & \underline{0.879} & \underline{0.033} & \underline{0.949} & \underline{0.890} & \underline{0.827} & \textbf{0.018} & \underline{0.948} & \underline{0.909} & \underline{0.881} & \underline{0.023} & \underline{0.955} \\
			SFGNet(Ours) & Semantics & CLIP+PVTv2-b5 & \textbf{0.906} & \textbf{0.880} & \textbf{0.032} & \textbf{0.951} & \textbf{0.891} & \textbf{0.829} & \textbf{0.018} & \textbf{0.949} & \textbf{0.912} & \textbf{0.884} & \textbf{0.021} & \textbf{0.957} \\
			\bottomrule
		\end{tabular}
	}
	\label{tab:comparison}
\end{table*}
}
\vspace{-0.5em}
\begin{figure*}[tb]
	\centerline{\includegraphics[width=\linewidth]{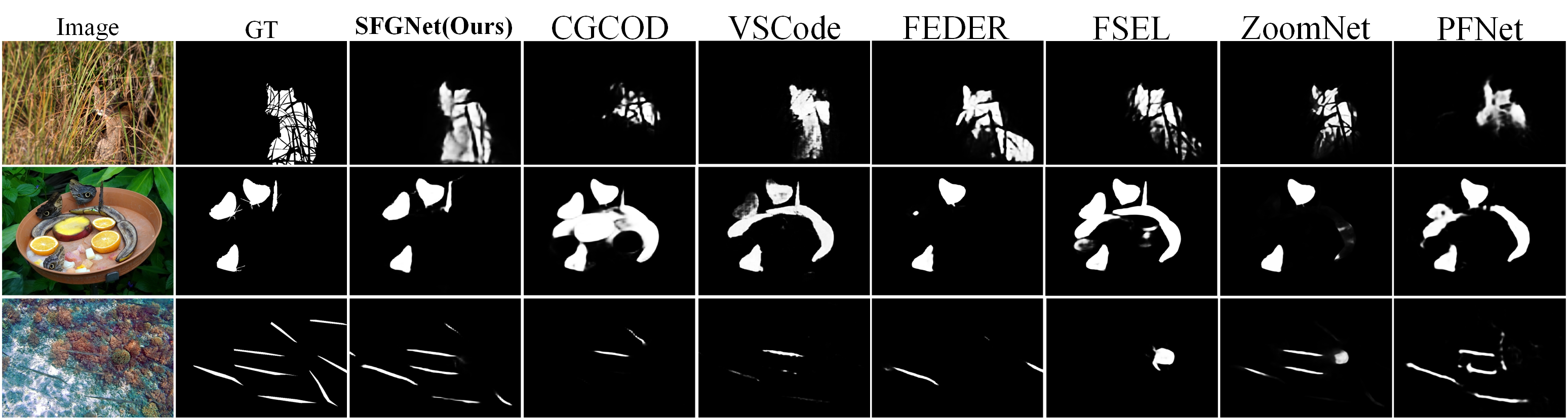}}
	\caption{Comparison of prediction results with state-of-the-art methods. SFGNet outperforms existing approaches in most cases.}
	\label{fig3}
\end{figure*}
\vspace{-0.5em}
\subsubsection{Evaluation Metrics}
\vspace{-0.3em}
We employed four widely used metrics to evaluate performance, including the S-measure ($S_m$)~\cite{Fan2017}, the weighted F-measure ($F_\beta^\omega$)~\cite{Margolin2014}, the mean absolute error (MAE), and the E-measure ($E_m$)~\cite{Fan2018e}.
\vspace{-1.2em}
\subsubsection{Implementation Details}
\vspace{-0.3em}
To reduce computational cost and leverage the pretrained knowledge of CLIP (ViT-L/336) \cite{Xu2023} and PVT (v2-b5), we kept the CLIP and PVT models frozen during training, while the parameters of other network components were learnable. We used AdamW to optimize the network parameters with a learning rate of 0.0001, a batch size of 12, and a total of 200 epochs. Various data augmentation strategies were applied to the training images, including random cropping, horizontal flipping, and color jittering. All experiments were conducted on an NVIDIA RTX 3090 GPU(24 GB memory).
\vspace{-1em}
\subsection{Comparison with State-of-the-art Methods}
We compared SFGNet with several recent and representative methods, including PFNet \cite{Mei2021}, ZoomNet \cite{Pang2020}, FSPNet \cite{Huang2023}, FEDER \cite{He2023}, HitNet \cite{Hu2023}, FSEL \cite{Sun2022}, CamoFormer \cite{Yin2024}, FHCNet \cite{Zhao2025}, VSCode \cite{Luo2023}, and CGCOD \cite{Zhang2024}. To ensure fair comparison, all prediction maps were either provided by the original authors or generated using the released codes. Table \ref{tab:comparison} reports the quantitative comparison between SFGNet and other models. Specifically, SFGNet outperforms the compared methods on CAMO, and NC4K datasets, while achieving comparable performance with CGCOD on COD10K. Qualitative comparisons are illustrated in Figure \ref{fig3}, which demonstrate that SFGNet performs better than state-of-the-art approaches in most cases, and is comparable in a few challenging scenarios. In addition, we evaluated different PVT backbones and found that pvt-v2-b5 achieves the best performance.
\vspace{-1em}
\subsection{Ablation Studies}
\vspace{-0.3em}
To evaluate the effectiveness of key components, we performed ablation studies on the NC4K dataset (4,121 images), as shown in Table~\ref{tab:ablation1}. The modules (BIN, MFA, ISEB, MBFM, and FSF) were progressively added to the baseline, yielding step-by-step performance gains and confirming the individual contribution of each. Visualization of intermediate features in Fig.~\ref{fig1} further illustrates the effectiveness of the proposed design.
{\renewcommand{\arraystretch}{0.85} 
\begin{table}[tb]
	\centering
	\caption{Ablation study of different components.}
	\resizebox{\linewidth}{!}{
		\begin{tabular}{l|cccc} 
			\toprule
			\multirow{2}{*}{Method} & \multicolumn{4}{c}{NC4K (4,121 images)} \\ 
			\cmidrule{2-5}
			& $S_m\uparrow$ & $F_\beta^\omega\uparrow$ & MAE$\downarrow$ & $E_m\uparrow$ \\ 
			\midrule
			Base                & 0.877          & 0.850          & 0.031          & 0.928 \\
			+BIN                & 0.876          & 0.852          & 0.032          & 0.929 \\
			+BIN+MFA            & 0.880          & 0.859          & 0.030          & 0.931 \\
			+BIN+MFA+ISEB       & 0.898          & 0.872          & 0.029          & 0.939 \\ 
			\midrule
			+BIN+MFA+MBFM       & 0.902          & 0.876          & 0.029          & 0.945 \\
			+BIN+MFA+MBFM+FSF   & 0.907          & 0.878          & 0.025          & 0.952 \\
			+BIN+MFA+MBFM+ISEB  & \underline{0.909}  & \underline{0.880}  & \underline{0.023}  & \underline{0.955} \\ 
			\midrule
			SFGNet (Ours)       & \textbf{0.912} & \textbf{0.884} & \textbf{0.021} & \textbf{0.957} \\
			\bottomrule
		\end{tabular}
	}
	\label{tab:ablation1}
\end{table}
}
\vspace{-1.3em}
\subsubsection{Analysis of different semantic prompts}
\vspace{-0.3em}
To evaluate the impact of different semantic prompts on SFGNet, we recorded the model performance on the COD10K dataset (2,026 images) with various prompts, as summarized in Table \ref{tab:ablation2}. The results show that the prompt “Camouflaged \textless class\textgreater~naturally blending into the surrounding environment.” achieves the best performance.
\vspace{-0.4em}
{\renewcommand{\arraystretch}{0.8} 
\begin{table}[t]
	\centering
	\caption{Impact of different semantic prompts on model performance.}
	\resizebox{\linewidth}{!}{
		\begin{tabular}{l|cccc} 
			\toprule
			\multirow{2}{*}{Semantic prompts} & \multicolumn{4}{c}{COD10K (2,026 images)} \\ 
			\cmidrule{2-5}
			& $S_m\uparrow$ & $F_\beta^\omega\uparrow$ & MAE$\downarrow$ & $E_m\uparrow$ \\ 
			\midrule
			"A photo of camouflaged object."                & 0.881 & 0.811 & 0.023 & 0.940 \\
			"A photo of camouflaged \textless class\textgreater." & 0.887 & \underline{0.824} & \underline{0.019} & 0.943 \\ 
			\midrule
			"Camouflaged object."                           & 0.867 & 0.785 & 0.027 & 0.927 \\
			"Camouflaged \textless class\textgreater."      & 0.850 & 0.755 & 0.028 & 0.912 \\ 
			\midrule
			"Camouflaged object naturally blending\\
			 into the surrounding environment." 
			& \underline{0.888} & \underline{0.824} & \underline{0.019} & \underline{0.946} \\
			\textbf{"Camouflaged \textless class\textgreater~}\\
			\textbf{naturally blending into the} \\
				\textbf{surrounding environment."} \textbf{(Ours)} 
			& \textbf{0.891} & \textbf{0.829} & \textbf{0.018} & \textbf{0.949} \\
			\bottomrule
		\end{tabular}
	}
	\label{tab:ablation2}
\end{table}
}
\vspace{-2em}
\section{Conclusion}
\label{sec:Conclusion}
\vspace{-1em}
In this paper, we proposed a Semantic and Frequency Guided Network (SFGNet) for camouflaged object detection. SFGNet integrates cross-modal semantic guidance and multi-band frequency modeling to combine high-level semantics with fine-grained structural cues. Specifically, semantic interaction modules (BIN, BCA, MFA) fuse multi-level visual and textual features, while frequency-domain modeling and spatial fusion (MBFM, FSF) enhance structural and boundary representations. The ISEB further improves result consistency. Extensive experiments show that SFGNet outperforms existing state-of-the-art methods on multiple benchmark datasets.
\begin{small}
\bibliographystyle{IEEEbib}
\bibliography{wdz2}
\end{small}

\end{document}